\documentclass{article}
\usepackage{spconf}

\usepackage{hyperref}       % hyperlinks
\usepackage{url}            % simple URL typesetting
\usepackage{booktabs}       % professional-quality tables

\usepackage{amsmath,amssymb,amsthm,amsbsy}
\usepackage{graphicx}
\usepackage{multirow}
\usepackage{algorithm}
\usepackage{algorithmic}
\usepackage{enumitem}
\usepackage{cite}

\theoremstyle{remark}
\newtheorem{remark}{Remark}

\newcommand{\round}{\operatorname{round}}

\title{Jointly Sparse Convolutional Neural Networks in Dual Spatial-Winograd Domains}
\name{Yoojin Choi, Mostafa El-Khamy, Jungwon Lee}
\address{SoC R\&D, Samsung Semiconductor Inc., San Diego, CA 92121, USA}
\begin{document}
%\ninept
%
\maketitle
\begin{abstract}
We consider the optimization of deep convolutional neural networks (CNNs) such that they provide good performance while having reduced complexity if deployed on either conventional systems with spatial-domain convolution or lower-complexity systems designed for Winograd convolution. The proposed framework produces one compressed model whose convolutional filters can be made sparse either in the spatial domain or in the Winograd domain. Hence, the compressed model can be deployed universally on any platform, without need for re-training on the deployed platform. To get a better compression ratio, the sparse model is compressed in the spatial domain that has a fewer number of parameters. From our experiments, we obtain $24.2\times$ and $47.7\times$ compressed models for ResNet-18 and AlexNet trained on the ImageNet dataset, while their computational cost is also reduced by $4.5\times$ and $5.1\times$, respectively.
\end{abstract}


\section{Introduction} \label{sec:intro}

Deep learning with convolutional neural networks (CNNs) has recently achieved performance breakthroughs in many of computer vision applications~\cite{lecun2015deep}. However, the large model size and huge computational complexity hinder the deployment of state-of-the-art CNNs on resource-limited platforms such as battery-powered mobile devices. Hence, it is of great interest to compress large-size CNNs into compact forms to lower their storage requirements and computational costs~\cite{sze2017efficient}.

CNN size compression has been actively investigated for memory and storage size reduction. Han et al.~\cite{han2015deep} showed impressive compression results by weight pruning, quantization using $k$-means clustering and Huffman coding. It has been followed by further analysis and mathematical optimization, and more efficient CNN compression schemes have been suggested afterwards, e.g., in \cite{choi2017towards,ullrich2017soft,agustsson2017soft,molchanov2017variational,louizos2017bayesian,choi2018universal,dai2018compressing}. Computational complexity reduction of CNNs has also been investigated on the other hand. The major computational cost of CNNs comes from the multiply-accumulate (MAC) operations in their convolutional layers~\cite[Table~II]{sze2017efficient}. There have been two directions to reduce the complexity of convolutions in CNNs:
\begin{itemize}[noitemsep,topsep=0em,leftmargin=1em]
\item First, instead of conventional spatial-domain convolution, it is proposed to use frequency-domain convolution~\cite{mathieu2013fast,vasilache2014fast} or Winograd convolution~\cite{lavin2016fast}. For typical small-size filters such as $3\times3$ filters, Lavin \& Gray~\cite{lavin2016fast} showed that Winograd convolution is more efficient than both spatial-domain convolution and frequency-domain convolution.
\item Second, weight pruning is another approach to reduce the number of MACs required for convolution by skipping the MACs involving pruned weights (zero weights). Previous work mostly focused on spatial-domain weight pruning to exploit sparse spatial-domain convolution of low complexity, e.g., see \cite{han2015deep,lebedev2016fast,wen2016learning,guo2016dynamic,lin2017runtime,park2017faster}. Recently, there have been some attempts to prune Winograd-domain weights~\cite{li2017enabling,liu2018efficient}.
\end{itemize}

\begin{figure*}
\centering
\includegraphics[width=.95\textwidth]{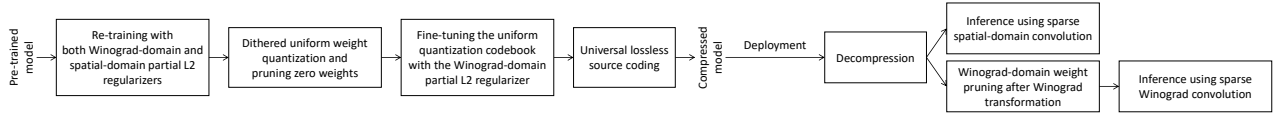}
\caption{Universal CNN weight pruning and compression for both sparse Winograd and sparse spatial-domain convolutions.\label{sec:intro:fig:01}}
\end{figure*}

Previous works either focused on spatial-domain weight pruning and compression or studied Winograd-domain weight pruning and complexity reduction. Compression of Winograd CNNs has never been addressed before. Other shortcomings of the previous works investigating the complexity reduction of Winograd CNNs are that their final CNNs are no longer backward compatible with spatial-domain convolution due to the non-invertibility of Winograd transformation, and hence they suffer from accuracy loss if they need to be run on the platforms that only support spatial-domain convolution. To our knowledge, this paper is the first to address the universal CNN pruning and compression framework for both Winograd and spatial-domain convolutions.

%Our proposed solutions are summarized in Figure~\ref{sec:intro:fig:01}. 
The main novelty of the proposed framework comes from the fact that it optimizes CNNs such their convolutional filters can be pruned either in the Winograd domain or in the spatial domain without accuracy loss and without extra training or fine-tuning in that domain. Our CNNs can be optimized for and compressed by universal quantization and universal source coding such that their decompressed convolutional filters still have sparsity in both Winograd and spatial domains. Hence, one universally compressed model can be deployed on any platform whether it uses spatial-domain convolution or Winograd convolution, and the sparsity of its convolutional filters can be utilized for complexity reduction in either domain. Since many low-power platforms, such as mobile phones, are expected to only support the inference of CNNs, and not their training, our approach is more desirable for wide-scale deployment of pre-trained models without worrying about underlying inference engines.

\section{Winograd convolution} \label{sec:winograd}

We review the Winograd convolution algorithm~\cite{winograd1980arithmetic} in this section. For the sake of illustration, consider that we are given a two-dimensional (2-D) input of size $H\times W$ and a 2-D filter of size $r\times r$ for convolution. For Winograd convolution, we first prepare a set of patches of size $n\times n$ extracted from the input with stride of $n-r+1\times n-r+1$ for $n\geq r$. Each of the $n\times n$ patches is convolved with the $r\times r$ filter by the Winograd convolution algorithm and produces an output patch of size $n-r+1\times n-r+1$.

Let $x$ and $y$ be one of the $n\times n$ input patches and its corresponding output patch, respectively, and let $w$ be the $r\times r$ filter. In Winograd convolution, the input and the filter are transformed into the Winograd domain by $X=FxF^T$ and $W=GwG^T$ using the Winograd transformation matrices~$F$ and $G$, respectively, where the superscript~$T$ denotes the matrix transpose. In the Winograd domain, both $X$ and $W$ are of size $n\times n$, and element-wise product of them follows. Then, the output is transformed back to the spatial domain by%using matrix~$S$ by
\begin{equation} \label{sec:winograd:eq:01}
y=S^T(W\odot X)S,
\end{equation}
where $\odot$ is the element-wise product of two matrices. The transformation matrices $F$, $G$, and $S$ are $(r,n)$-specific and can be obtained from the Chinese remainder theorem (e.g., see \cite[Section~5.3]{blahut2010fast}). For more details, see \cite[Section~4]{lavin2016fast}.%In case of $C$ input channels, the inverse transformation in \eqref{sec:winograd:eq:01} can be deployed once after summation over all channels of the element-wise product outputs in the Winograd domain (see \cite[Section~4]{lavin2016fast}).

\section{Training with joint sparsity constraints}

In this section, we present our CNN training method with regularization for joint spatial-Winograd sparsity constraints. %, to enable efficient deployment of pre-trained CNNs in either domain, without additional training for deployment.
%\subsection{CNN model}
We consider a typical CNN model consisting of $L$ convolutional layers. The input of layer~$l$ has $C_l$ channels of size $H_l\times W_l$ and the output has $D_l$ channels of size $H_l-r_l+1\times W_l-r_l+1$, where the input is convolved with $D_l$ filters of size $r_l\times r_l\times C_l$. For $1\leq l\leq L$, $1\leq i\leq C_l$ and $1\leq j\leq D_l$, %let $x_l(i)$ and $y_l(j)$ be the 2-D input and output feature maps of channel~$i$ and channel~$j$, respectively, for layer~$l$, and
let $w_l(i,j)$ be the 2-D convolutional filter for input channel~$i$ and output channel~$j$ of layer~$l$. %The convolutional layer~$l$ produces
%\[
%y_l(j)=\sum_{i=1}^{C_l}w_l(i,j)*x_l(i),
%\]
%for $1\leq j\leq D_l$, where $*$ is 2-D spatial-domain convolution.

\subsection{Regularization for jointly sparse convolutional filters} \label{sec:reg}

%In this subsection, we introduce our Winograd-domain and spatial-domain partial L2 regularizers to attain convolutional filters that are sparse in both the Winograd domain and the spatial domain. 
We choose L2 regularizers to promote sparsity, although other regularizers such as L1 regularizers can be used instead (see Remark~\ref{remark:02} for more discussion). Let $\mathbf{w}$ be the set of all convolutional filters of $L$ layers, which are learnable, i.e., $\mathbf{w}\equiv\{w_l(i,j),1\leq l\leq L,1\leq i\leq C_l,1\leq j\leq D_l\}$. Moreover, given any matrix~$A$, we define $1_{|A|\leq\theta}$ as the matrix that has the same size as $A$ while its element is one if the corresponding element~$a$ in $A$ satisfies $|a|\leq\theta$ and is zero otherwise.

%\textbf{Winograd-domain partial L2 regularization}:
To optimize CNNs under Winograd-domain sparsity constraints, we introduce the Winograd-domain partial L2 regularizer given by
\begin{multline} \label{sec:reg:eq:01}
R_{\text{WD}}(\mathbf{w};s_{\text{WD}})
=\frac{1}{N_{\text{WD}}}\sum_{l=1}^L\sum_{i=1}^{C_l}\sum_{j=1}^{D_l} \\
\|(G_lw_l(i,j)G_l^T)\odot1_{|G_lw_l(i,j)G_l^T|\leq\theta_{\text{WD}}(s_{\text{WD}})}\|^2,
\end{multline}
where $\|\cdot\|$ denotes the L2 norm and $G_l$ is the Winograd transformation matrix determined by the filter size and the input patch size of layer~$l$ (see Section~\ref{sec:winograd}); $N_{\text{WD}}$ is the total number of Winograd-domain weights of all $L$ layers.

The L2 regularization in \eqref{sec:reg:eq:01} is applied only to a part of Winograd-domain weights if their magnitude values are not greater than the threshold value~$\theta_{\text{WD}}(s_{\text{WD}})$. Although the constraints are on the Winograd-domain weights, they translate as the constraints on the corresponding spatial-domain weights, and the optimization is done in the spatial domain. %; this facilitates the optimization with the additional sparsity constraints in the spatial domain as will be clarified below.
Due to the Winograd-domain partial L2 regularization, spatial-domain weights are updated towards the direction to yield diminishing Winograd-domain weights in part after training and being transformed into the Winograd domain.

Given a desired sparsity level~$s_{\text{WD}}$ (\%) in the Winograd domain, we set the threshold value~$\theta_{\text{WD}}(s_{\text{WD}})$ to be the $s_{\text{WD}}$-th percentile of Winograd-domain weight magnitude values. The threshold is updated at every training iteration as weights are updated, and it decreases as training goes on since the regularized Winograd-domain weights within the $s_{\text{WD}}$-th percentile converge to small values. After finishing the regularized training, we finally have a subset of Winograd-domain weights clustered very near zero, which can be pruned (i.e., set to zero) at minimal accuracy loss (see Figure~\ref{sec:train:fig:01}).
%In particular, as the regularized Winograd-domain weights become smaller in magnitude, the threshold value also decreases accordingly following their $s_{\text{WD}}$-th percentile magnitude value. We can use a separate threshold value for each layer and control the sparsity of each layer, but here we use one common threshold value for all layers for simplicity.

%\textbf{Spatial-domain partial L2 regularization}:
To optimize CNNs while having sparsity in the spatial domain, similar to \eqref{sec:reg:eq:01}, we regularize the cost function by the partial sum of L2 norms of spatial-domain weights as follows:%, determined by $\theta_{\text{SD}}(s_{\text{SD}})$ given a target sparsity level~$s_{\text{SD}}$ (\%), as below:
\begin{multline} \label{sec:reg:eq:02}
R_{\text{SD}}(\mathbf{w};s_{\text{SD}})
=\frac{1}{N_{\text{SD}}}\sum_{l=1}^L\sum_{i=1}^{C_l}\sum_{j=1}^{D_l} \\
\|w_l(i,j)\odot1_{|w_l(i,j)|\leq\theta_{\text{SD}}(s_{\text{SD}})}\|^2,
\end{multline}
where $N_{\text{SD}}$ is the total number of spatial-domain weights of all $L$ layers, and $\theta_{\text{SD}}(s_{\text{SD}})$ is the threshold given a target sparsity level~$s_{\text{SD}}$ (\%) for spatial-domain weights.

\subsection{Training with learnable regularization coefficients} \label{sec:train}

Combining the regularizers in \eqref{sec:reg:eq:01} and \eqref{sec:reg:eq:02}, the cost function~$C$ to minimize in training is given by
\begin{multline} \label{sec:reg:eq:03}
C(\mathcal{X};\mathbf{w})
=E(\mathcal{X};\mathbf{w})
+\lambda_{\text{WD}} R_{\text{WD}}(\mathbf{w};s_{\text{WD}}) \\
+\lambda_{\text{SD}} R_{\text{SD}}(\mathbf{w};s_{\text{SD}}),
\end{multline}
for $\lambda_{\text{WD}}>0$ and $\lambda_{\text{SD}}>0$, where $\mathcal{X}$ is the training dataset and the $E$ is the network loss function such as the cross-entropy loss for classification or the mean-squared-error loss for regression. %We emphasize that training is performed in the spatial domain with conventional spatial-domain convolution and we update spatial-domain filters in $\mathbf{w}$.%, while the regularizers steer the filters to have a desired percentage of weights with small or near-zero values either in the spatial domain or in the Winograd domain when transformed, which are safe to prune at little accuracy loss.
Here, we introduce two regularization coefficients~$\lambda_{\text{SD}}$ and $\lambda_{\text{WD}}$. Conventionally, we use a fixed value for a regularization coefficient. However, we observe that using fixed regularization coefficients for the whole training is not efficient to find good sparse models. For small coefficients, regularization is weak and we cannot reach the desired sparsity after training.
%Hence, a significant accuracy loss may follow when we actually prune weights for sparse convolution.
For large coefficients, on the other hand, we can achieve the desired sparsity, but it likely comes with considerable accuracy loss due to strong regularization.

\setlength{\tabcolsep}{0.1em}
\begin{figure}
\centering
{\scriptsize
\begin{tabular}{cccc}
 & (a) Iterations=0 & (b) Iterations=100k & (c) Iterations=200k \\
\raisebox{1.5em}{\rotatebox{90}{\shortstack[c]{Winograd domain\\~\\~}}} &
\includegraphics[width=0.14\textwidth]{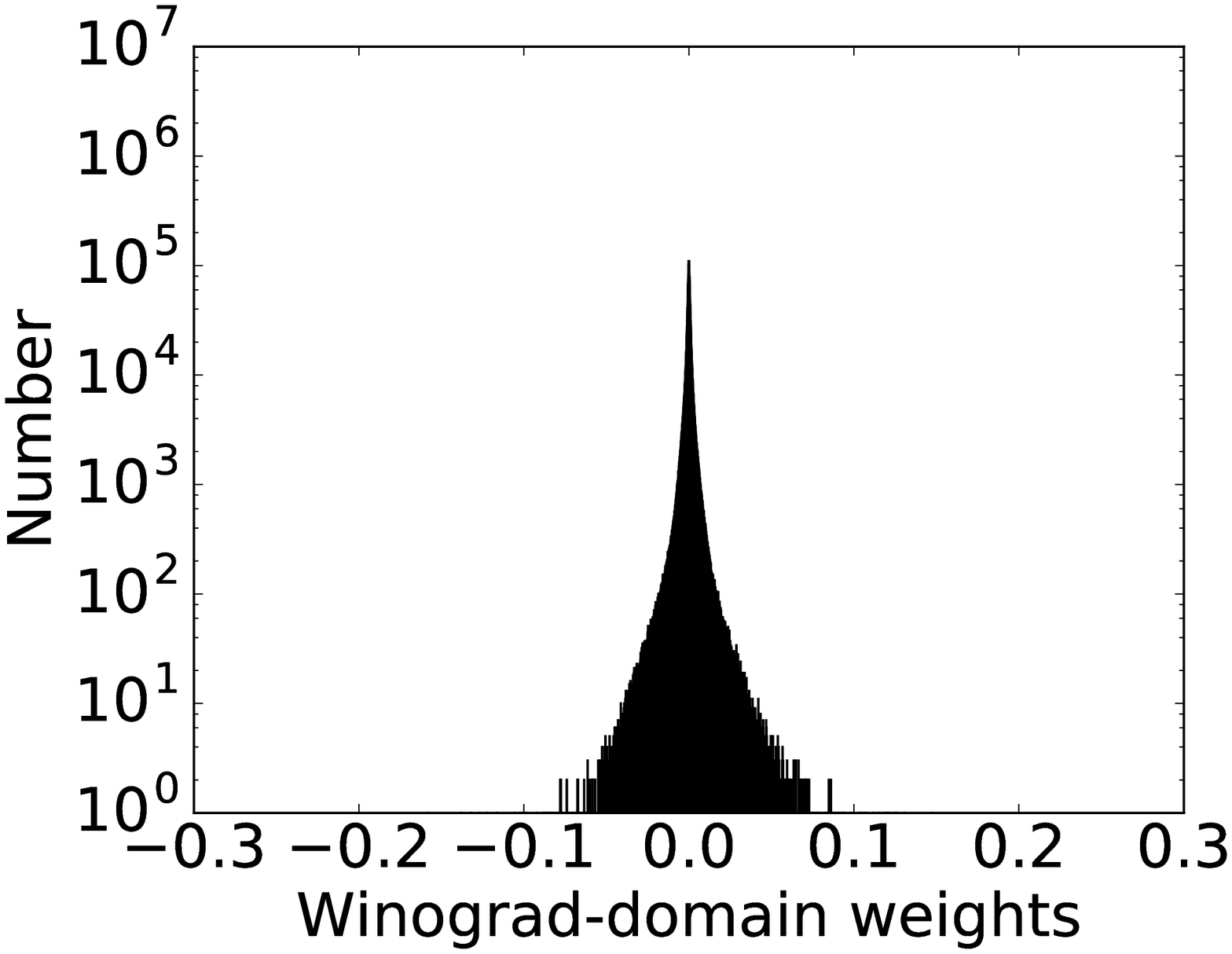} &
\includegraphics[width=0.14\textwidth]{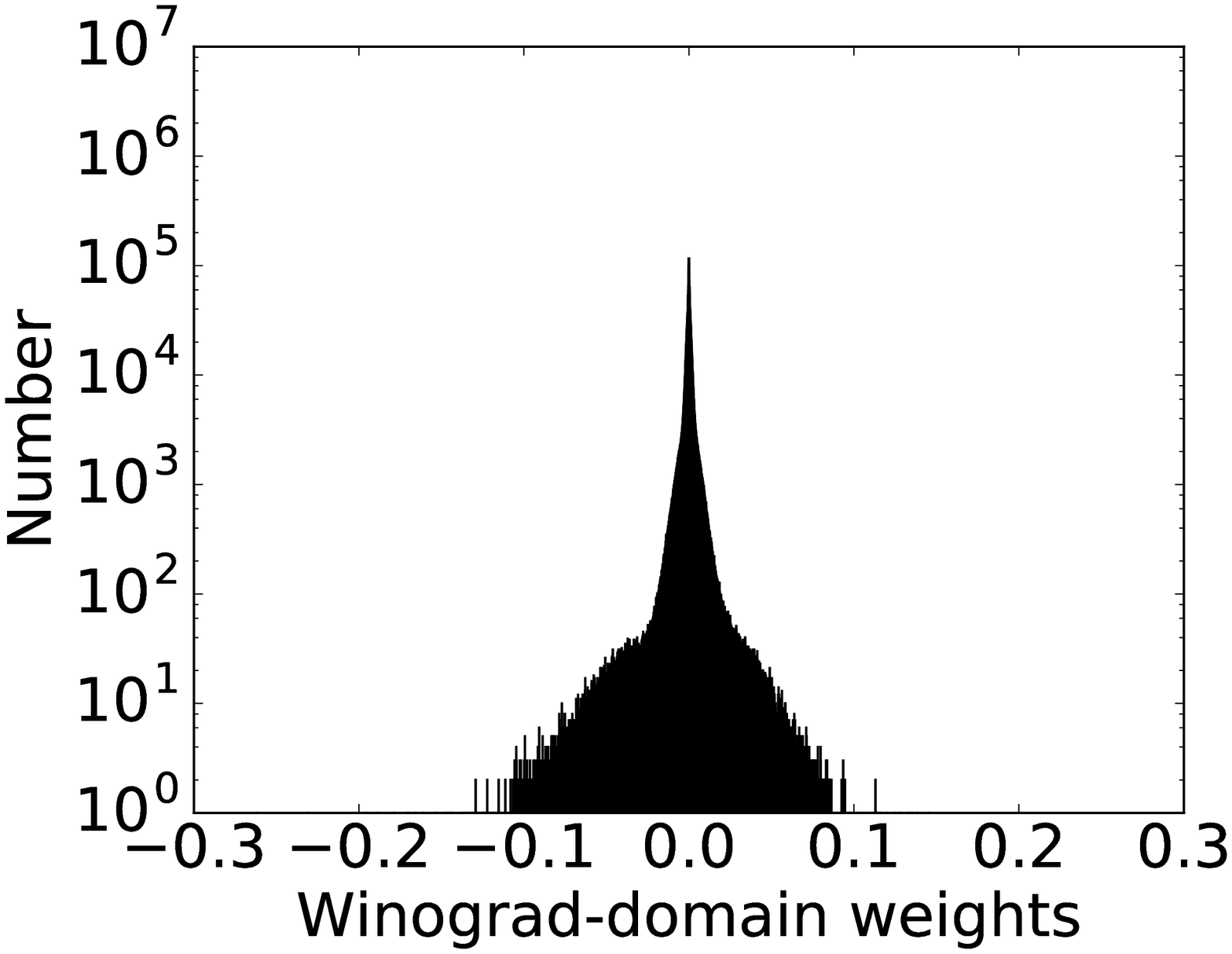} &
\includegraphics[width=0.14\textwidth]{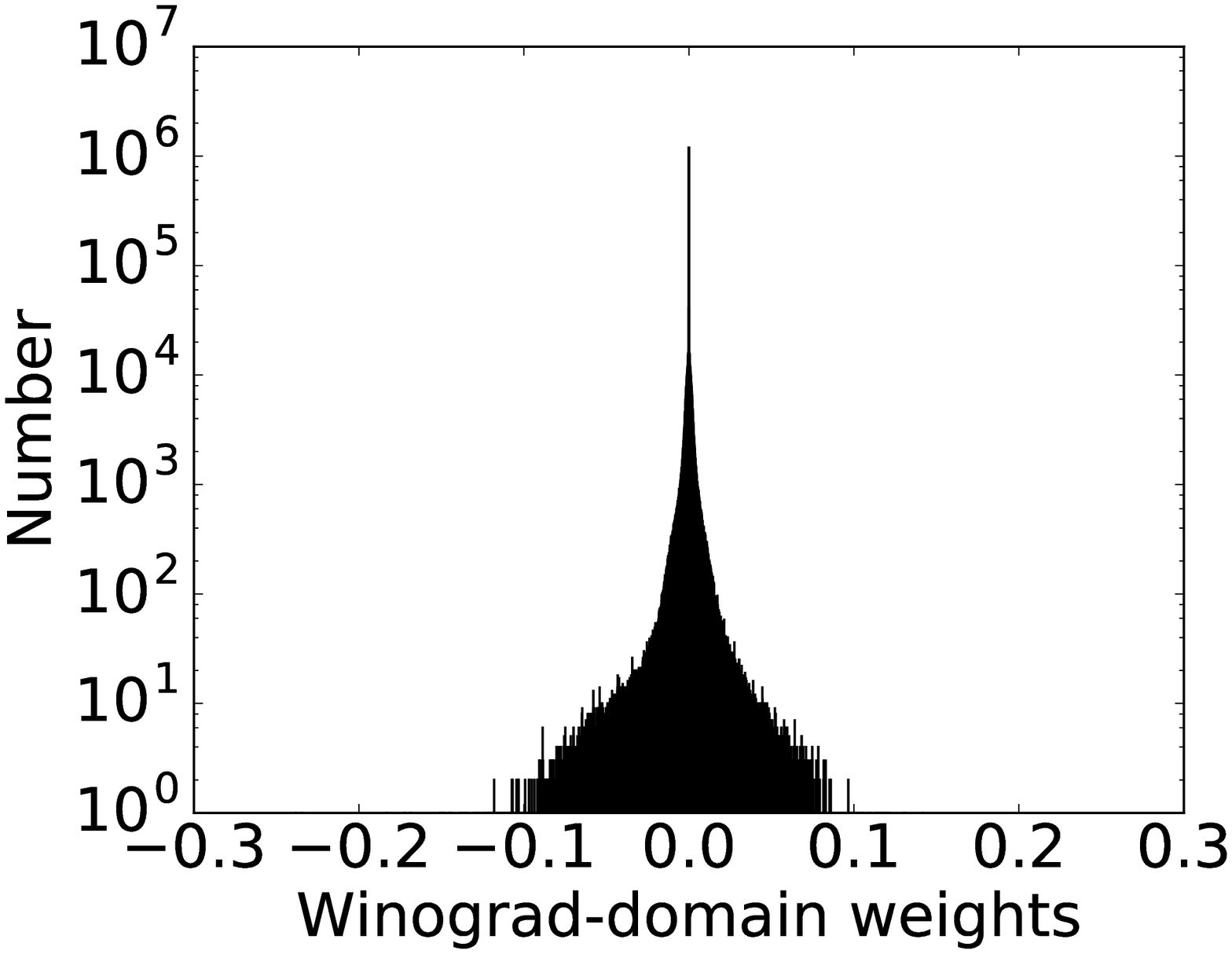} \\
\raisebox{1.5em}{\rotatebox{90}{\shortstack[c]{Spatial domain\\~\\~}}} &
\includegraphics[width=0.14\textwidth]{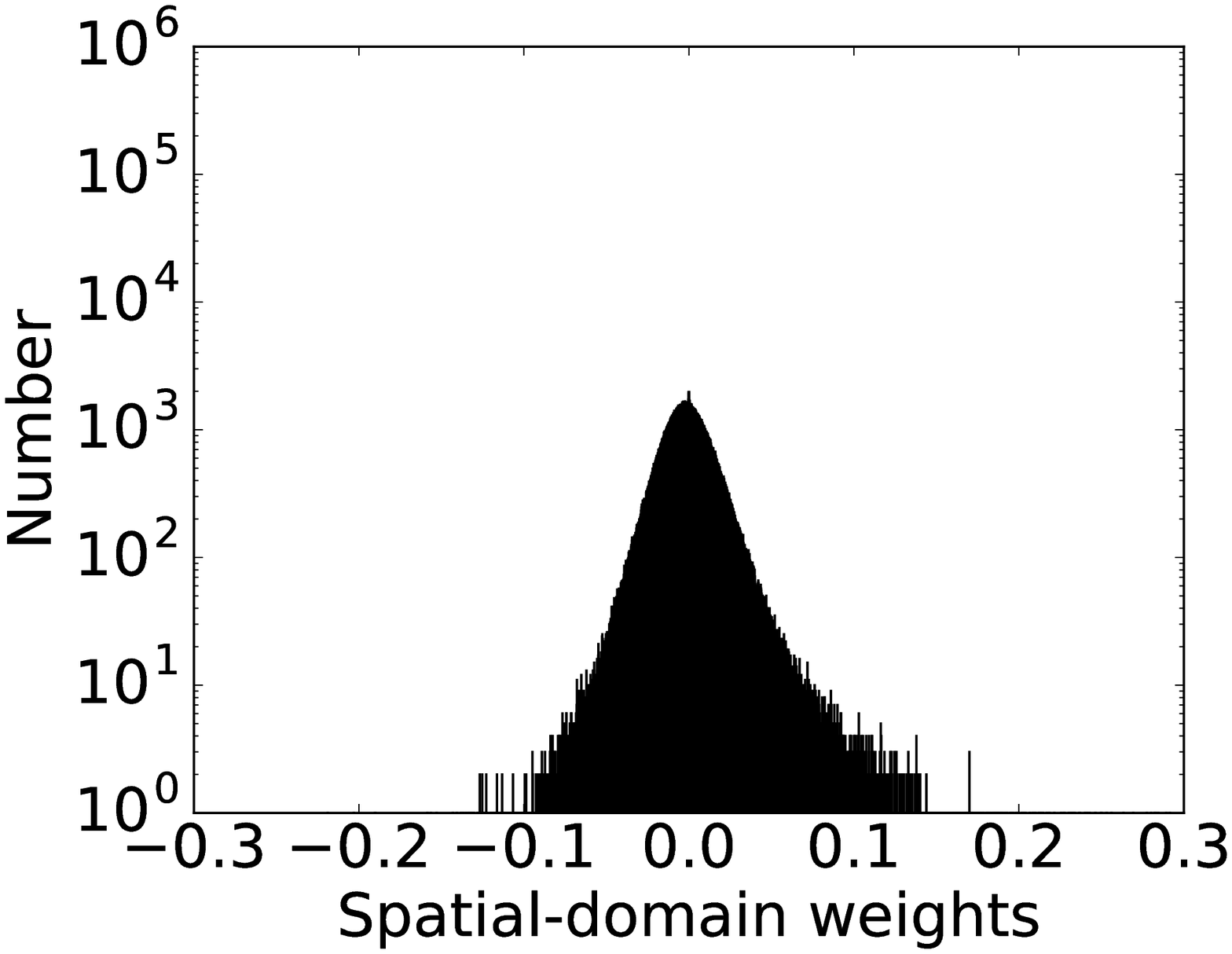} &
\includegraphics[width=0.14\textwidth]{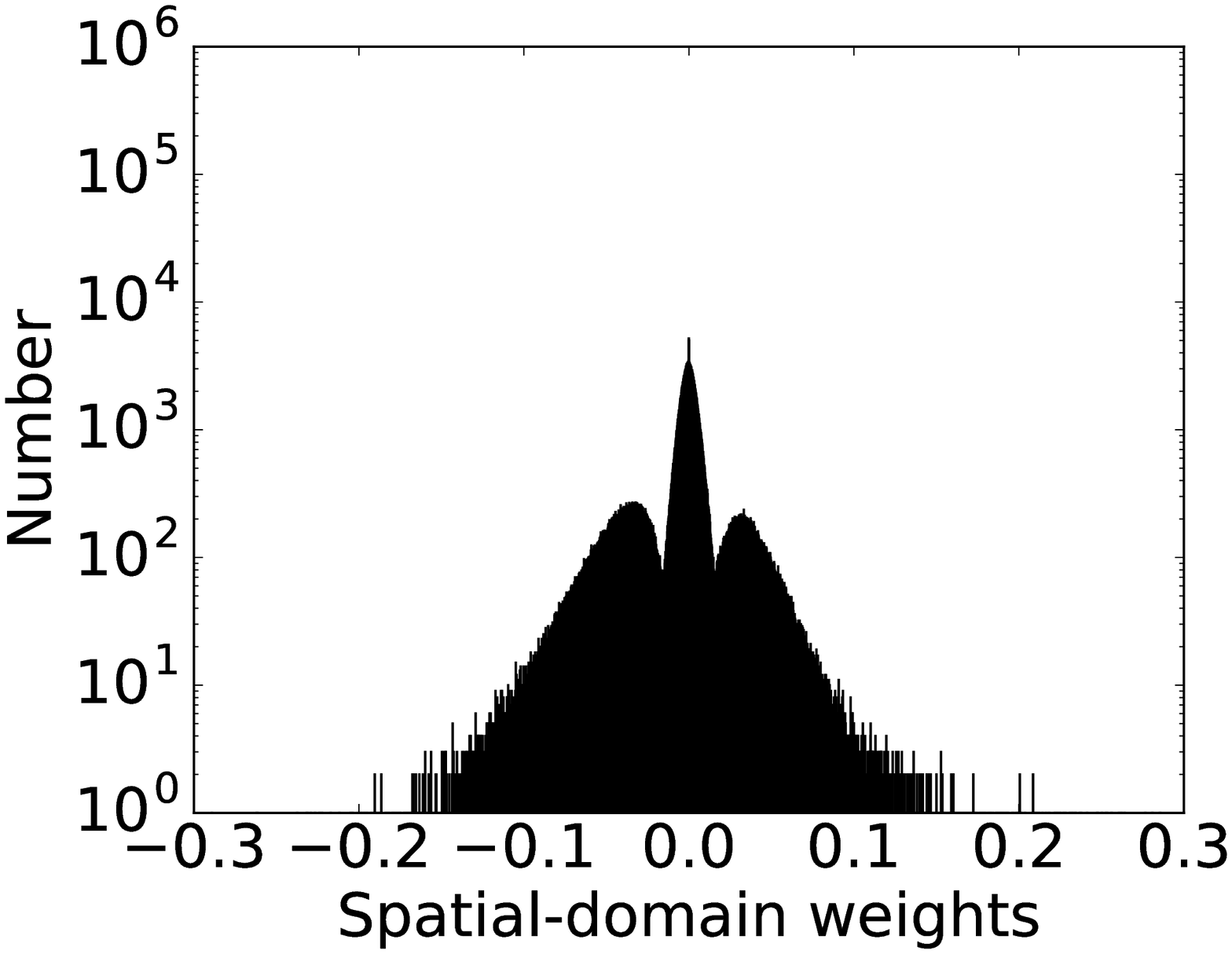} &
\includegraphics[width=0.14\textwidth]{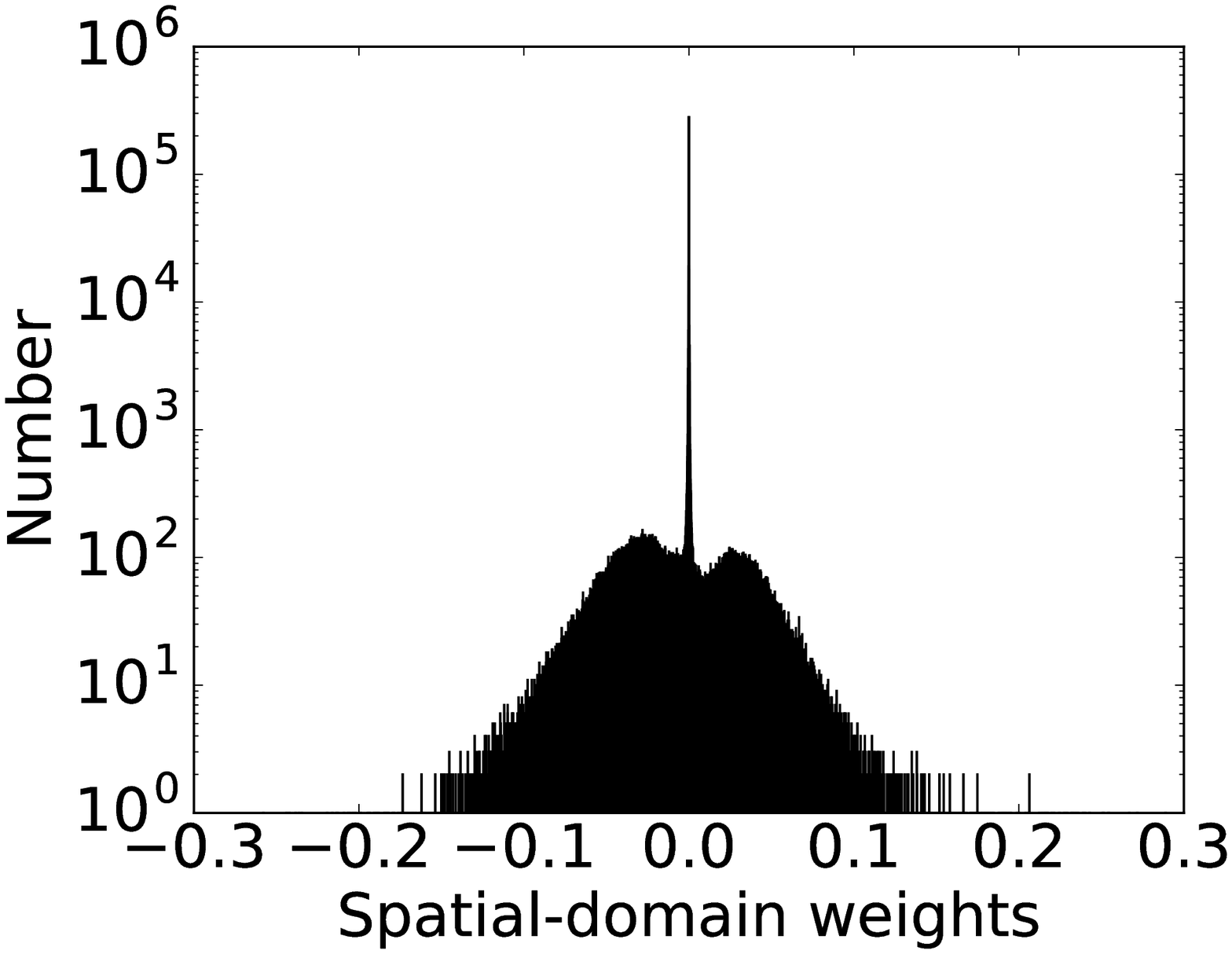} \\
\end{tabular}
}
\caption{Weight histogram snapshots of the AlexNet second convolutional layer.\label{sec:train:fig:01}}
\end{figure}

%\textbf{Learnable regularization coefficient}:
To overcome the problems with fixed regularization coefficients, we propose novel \emph{learnable regularization coefficients}, i.e., we let regularization coefficients be learnable parameters. Starting from a small initial coefficient value, we learn an accurate model with little regularization. As training goes on, we induce the regularization coefficients to increase gradually so that the performance does not degrade much but we finally have sparse convolutional filters at the end of training. To this end, we replace $\lambda_{\text{WD}}$ and $\lambda_{\text{SD}}$ with $e^{\zeta_{\text{WD}}}$ and $e^{\zeta_{\text{SD}}}$, respectively, and learn $\zeta_{\text{WD}}$ and $\zeta_{\text{SD}}$ instead, for the sake of guaranteeing that the regularization coefficients always positive in training. Then, we include an additional regularization term, i.e., $-\alpha(\zeta_{WD}+\zeta_{SD})$ for $\alpha>0$, which penalizes small regularization coefficients and encourages them to increase in training. As a result, the cost function in \eqref{sec:reg:eq:03} is altered into
\begin{multline} \label{sec:reg:eq:04}
C(\mathcal{X};\mathbf{w},\zeta_{\text{WD}},\zeta_{\text{SD}})
=E(\mathcal{X};\mathbf{w})
+e^{\zeta_{\text{WD}}}R_{\text{WD}}(\mathbf{w};s_{\text{WD}}) \\
+e^{\zeta_{\text{SD}}}R_{\text{SD}}(\mathbf{w};s_{\text{SD}})
-\alpha(\zeta_{\text{WD}}+\zeta_{\text{SD}}).
\end{multline}
%Observe that we introduced a new hyper-parameter~$\alpha$, while making regularization coefficients learnable. The trade-off between the loss and the regularization is now controlled by the new hyper-parameter~$\alpha$ instead of regularization coefficients, which is beneficial since $\alpha$ is not directly related to either of the loss or the regularization, and we can induce smooth transition to a sparse model.% and the optimal trade-off between them can be learned.
The indicator functions in \eqref{sec:reg:eq:01} and \eqref{sec:reg:eq:02} are non-differentiable, which is however not a problem when computing the derivatives of \eqref{sec:reg:eq:04} in practice for stochastic gradient descent.

\begin{figure}
\centering
{\scriptsize
\begin{tabular}{cc}
\includegraphics[height=0.06\textwidth]{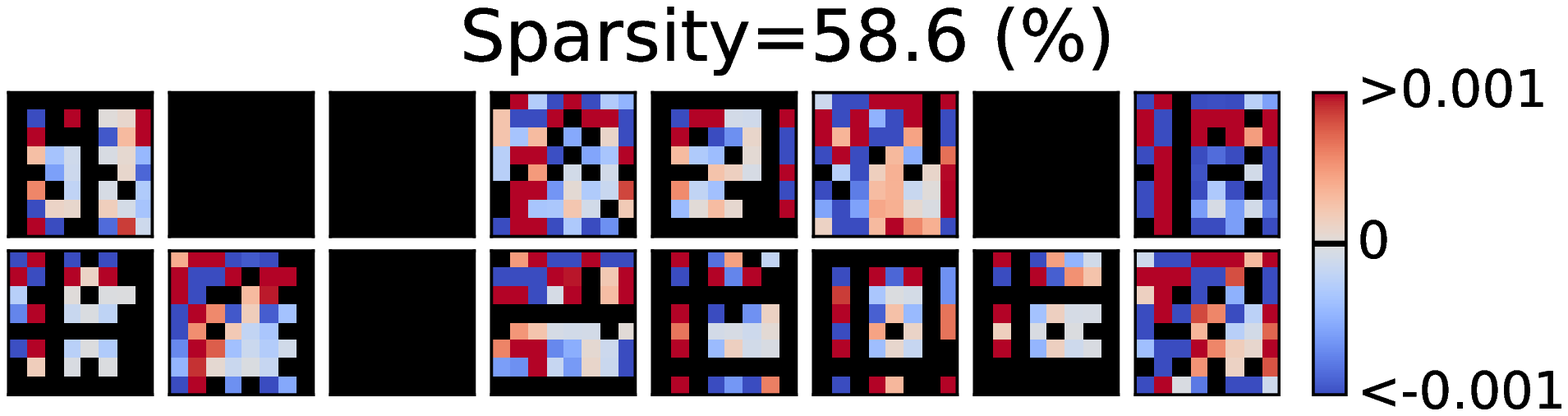} &
\includegraphics[height=0.06\textwidth]{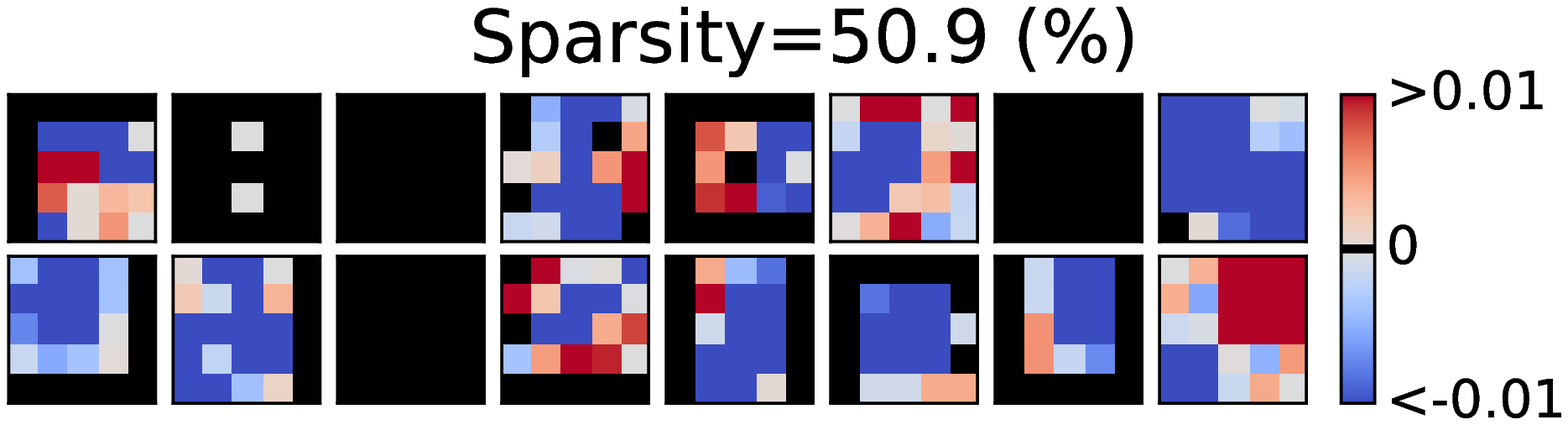} \\
(a) Winograd domain & (b) Spatial domain \\
\end{tabular}
}
\caption{Sparse convolutional filters from the AlexNet second convolutional layer, obtained after pruning in either domain.\label{sec:train:fig:03}}
\end{figure}

%\textbf{Evolution of weight histogram}:
In Figure~\ref{sec:train:fig:01}, we present how the weight histogram (distribution) of the AlexNet second convolutional layer evolves in the Winograd and spatial domains when trained with the proposed cost function in \eqref{sec:reg:eq:04}. %The plots show weight histogram snapshots captured in the Winograd domain and in the spatial domain at the second convolutional layer of the AlexNet model~\cite{krizhevsky2012imagenet}.
Observe that a part of the weights converges to zero in both domains, %and we finally have a peak at zero,
which can be pruned at minimal accuracy loss. %, in each domain.
%
%\textbf{Examples of pruned filters}:
In Figure~\ref{sec:train:fig:03}, we present sparse convolutional filters obtained after pruning either in the Winograd domain and in the spatial domain. They are sampled from the $5\times5$ filters of the AlexNet second convolutional layer, where we use Winograd convolution of $(r,n)=(5,8)$ in Section~\ref{sec:winograd}.

%\begin{figure}
%\centering
%\includegraphics[width=\columnwidth]{fig/fig01}
%\caption{Sparse model obtained after pruning either in the spatial domain or in the Winograd domain.\label{sec:jointsparsity:fig:01}}
%\end{figure}

\begin{remark} \label{remark:02}
As observed above, we have presented our algorithms using L2 regularizers. Often L1 norms are used to promote sparsity (e.g., see \cite{chen2001atomic}), but here we suggest using L2 instead, since our goal is to induce small-value weights rather than to drive them to be really zero. The model re-trained with our L2 regularizers is still dense and not sparse before pruning. However, it is jointly regularized to have many small-value weights, which can be pruned at negligible loss, in both domains. The sparsity is actually attained only after pruning its small-value weights in either domain. %, i.e., in any of the spatial or the Winograd domain.
This is to avoid the fundamental limit of joint sparsity, similar to the uncertainty principle of the Fourier transform~\cite{donoho1989uncertainty}.
\end{remark}

\section{Universal compression and dual domain deployment} \label{sec:univ}

A universal CNN compression framework is proposed in \cite{choi2018universal}, where CNNs are optimized for and compressed by universal quantization~\cite{ziv1985universal} and universal entropy source coding with schemes such as the variants of Lempel--Ziv--Welch~\cite{ziv1977universal,ziv1978compression,welch1984technique} and the Burrows--Wheeler transform~\cite{effros2002universal}. %Of particular interest for universal quantization is randomized uniform quantization, where uniform random dithering makes the distortion independent of the source, and the gap of its rate from the rate-distortion bound at any distortion level is provably no more than $0.754$ bits per sample for any source~\cite{zamir1992universal}. Universal compression has practical advantages as it is easily applicable to any CNN model at any desired compression rate, without the extra burden required by previous approaches to compute or estimate the statistics of the CNN weights, and is guaranteed to perform near-optimally.
%We compress the jointly sparse CNN model from Section~\ref{sec:train} by universal compression in the spatial domain for universal deployment.

Our universal compression pipeline under joint sparsity constraints is summarized in Figure~\ref{sec:intro:fig:01}. We randomize spatial-domain weights by adding uniform random dithers, and quantize the dithered weights uniformly with interval~$\Delta$ by
\begin{equation} \label{sec:comp:eq:01}
q_i=\Delta\cdot\round((a_i+U_i)/\Delta),
\end{equation}
where $a_1,\dots,a_{N_{\text{SD}}}$ are the individual spatial-domain weights of all $L$ layers, and $U_1,\dots,U_{N_{\text{SD}}}$ are independent and identically distributed uniform random variables with the support of $[-\Delta/2,\Delta/2]$; the rounding yields the closest integer of the input. The weights rounded to zero in \eqref{sec:comp:eq:01} are pruned and fixed to be zero in the compressed model. The random dithering values or their random seed are assumed to be known at deployment, and the dithering values are cancelled for the unpruned weights after decompression by $\hat{q}_i=q_i-U_i\cdot1_{q_i\neq0}$, 
%\begin{equation} \label{sec:comp:eq:02}
%\hat{q}_i
%=q_i-U_i\cdot1_{q_i\neq0},%=\Delta\cdot\round((a_i+U_i)/\Delta)-U_i\cdot1_{q_i\neq0},
%\end{equation}
where $\hat{q}_i$ is the final deployed value of weight~$a_i$ for inference. 

For one compressed model, we make the model actually sparse in the spatial domain by pruning small-value weights that are quantized to zero in the spatial domain. The resulting quantized model is sparse in the spatial domain, but it becomes dense in the Winograd domain. To recover the sparsity in the Winograd domain and to compensate the accuracy loss from quantization, we fine-tune the spatial-domain quantization codebook with the Winograd-domain L2 regularizer. Using the cost function~$C=E+e^{\zeta_{\text{WD}}}R_{\text{WD}}-\alpha\zeta_{\text{WD}}$ instead of \eqref{sec:reg:eq:04}, 
%\[
%%C=E+e^{\zeta_{\text{WD}}}R_{\text{WD}}-\alpha\zeta_{\text{WD}},
%C(\mathcal{X};\mathbf{w},\zeta_{\text{WD}})
%=E(\mathcal{X};\mathbf{w})
%+e^{\zeta_{\text{WD}}}R_{\text{WD}}(\mathbf{w};s_{\text{WD}})
%-\alpha\zeta_{\text{WD}}.
%\]
%In fine-tuning, the pruned weights (i.e., weights quantized to zero) stay zero, while non-zero quantized values are fine-tuned. 
the average gradient is computed for unpruned weights that are quantized to the same value in \eqref{sec:comp:eq:01}. Then, their shared quantized value in the codebook is updated by gradient descent using the average gradient of them.
%, which is given by
%\begin{equation} \label{sec:comp:eq:03}
%c_n(t)
%=c_n(t-1)-\eta\frac{1}{|\mathcal{I}_n|}\sum_{i\in\mathcal{I}_n}\nabla_{a_i}C(t-1),\ \ \ n\neq0,
%\end{equation}
%where $t$ is the iteration time, $\eta$ is the learning rate, and $\mathcal{I}_n$ is the index set of all weights that are quantized to the same value~$c_n=n\Delta$ in \eqref{sec:comp:eq:01} for some non-zero integer~$n$.
%After the codebook is updated, individual weights are updated by following their shared quantized value in the codebook, i.e., %$\hat{q}_i(t)=c_n(t)-U_i$ for all $i\in\mathcal{I}_n$ and $n\neq0$.
%\[
%\hat{q}_i(t)=c_n(t)-U_i, \ \ \ \forall i\in\mathcal{I}_n,\ \ \ n\neq0.
%\]
We emphasize here that the pruned weights in \eqref{sec:comp:eq:01} are not fine-tuned and stay zero. 

At deployment, the compressed model is decompressed to get unpruned spatial-domain weights. Then, the CNN can be deployed in the spatial domain with the desired sparsity. If we deploy the CNN in the Winograd domain, its convolutional filters are transformed into the Winograd domain, and pruned to the desired sparsity level (see deployment in Figure~\ref{sec:intro:fig:01}).

\section{Experiments} \label{sec:exp}

\begin{figure}[t]
\centering
{\scriptsize
\begin{tabular}{cc}
\includegraphics[height=0.11\textwidth]{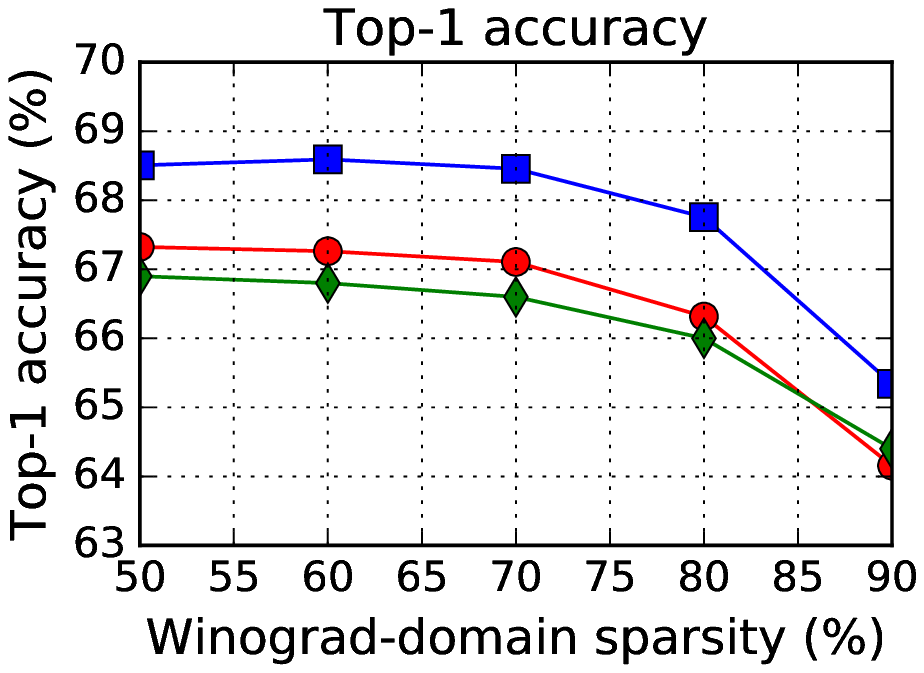} &
\includegraphics[height=0.11\textwidth]{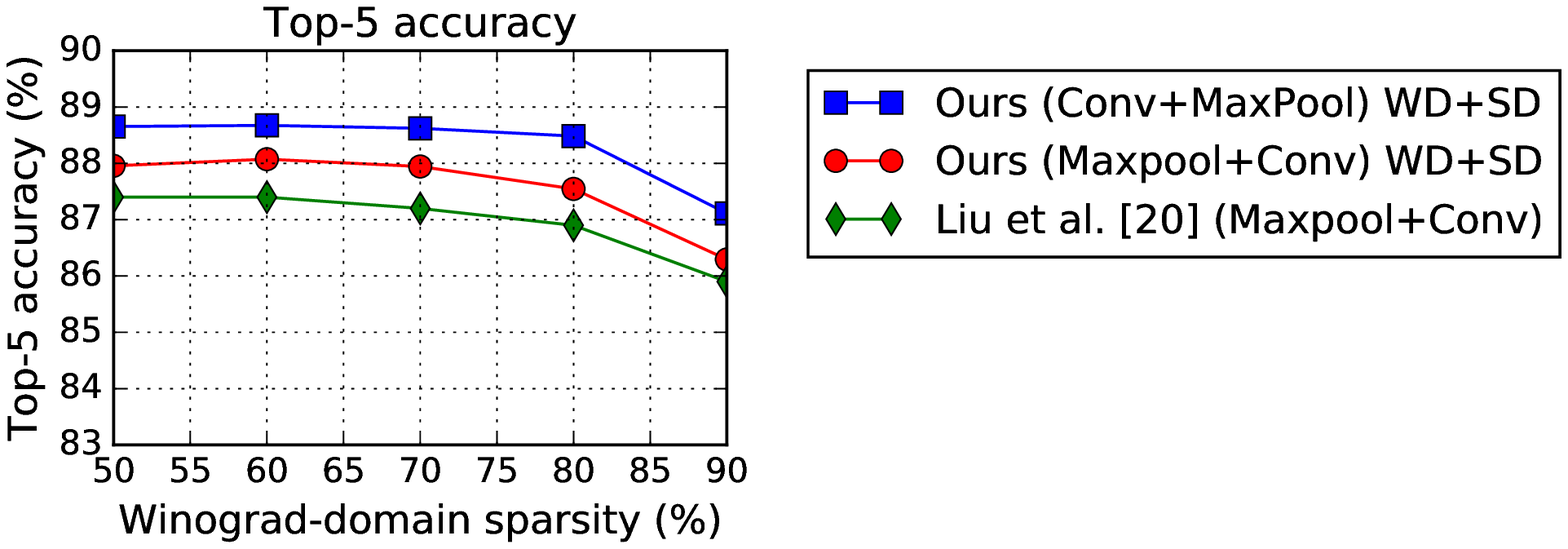} \\
\end{tabular}
}
\caption{Accuracy comparison for the pruned ResNet-18 models at different sparsity levels in the Winograd domain.\label{sec:exp:resnet18:fig:01}}
\end{figure}

We experiment with our universal CNN pruning and compression scheme on the ResNet-18 model~\cite{he2016deep} trained for the ImageNet ILSVRC 2012 dataset~\cite{russakovsky2015imagenet}. As in \cite{liu2018efficient}, we modify the original ResNet-18 model by replacing its convolutional layers of stride $2\times2$ with convolutional layers of stride $1\times1$ and max-pooling layers, to deploy Winograd convolution for all possible convolutional layers. One difference from \cite{liu2018efficient} is that we place max-pooling after convolution (Conv+Maxpool) instead of placing it before convolution (Maxpool+Conv). Our modification provides better accuracy (see Figure~\ref{sec:exp:resnet18:fig:01}).% although it comes with more computations.

\begin{table}[t]
\caption{Accuracy and complexity of pruned ResNet-18 models when using different regularization methods.\label{sec:exp:resnet18:tbl:01}}\vspace{.5em}
\centering
{\scriptsize
\begin{tabular}{ccccccc}
\toprule
\multirow{2}{*}{\shortstack[c]{Regularization\\(sparsity~$s$)}}
 & \multirow{2}{*}{\shortstack[c]{Pruning\\ratio}}
 & \multicolumn{2}{c}{(1) Spatial domain} & \multicolumn{2}{c}{(2) Winograd domain} \\
\cmidrule(l{2pt}r{2pt}){3-4} \cmidrule(l{2pt}r{2pt}){5-6}
 & & Top-1 / Top-5 & \# MACs   & Top-1 / Top-5 & \# MACs   \\
 & & accuracy      & per image & accuracy      & per image \\
\midrule
Pre-trained model
               & -    & 68.2 / 88.6 & 2347.1M & 68.2 / 88.6 & 1174.0M \\
\midrule                                                                     
SD (80\%)      & 80\% & 67.8 / 88.4 & 837.9M  & 56.9 / 80.7 & 467.0M  \\
WD (80\%)      & 80\% & 44.0 / 70.5 & 819.7M  & 68.4 / 88.6 & 461.9M  \\
WD+SD (80\%)   & 80\% & 67.8 / 88.5 & 914.9M  & 67.8 / 88.5 & 522.6M  \\
\bottomrule
\end{tabular}
}
\end{table}

The Winograd-domain regularizer is applied to all $3\times3$ convolutional filters. We assume to use Winograd convolution of $(r,n)=(3,4)$ for $3\times3$ filters (see Section~\ref{sec:winograd}). The spatial-domain regularizer is applied to all convolutional and fully-connected layers not only for pruning but also for compression in the spatial domain. We use the Adam optimizer~\cite{kingma2014adam} with the learning rate of $1\text{e-}5$ for $500k$ iterations with the batch size of $128$. We set $\alpha=1$ in \eqref{sec:reg:eq:04}. %The initial values for $\zeta_{\text{WD}}$ and $\zeta_{\text{SD}}$ are both set to be $10$, and they are updated using the Adam optimizer with the learning rate of $1\text{e-}4$.

In Table~\ref{sec:exp:resnet18:tbl:01}, we summarize the accuracy and the number of MACs to process one input image for pruned ResNet-18 models.
%The number of MACs for Winograd convolution is counted by following \cite[Section~5]{lavin2016fast}.
We compare three models obtained with spatial-domain regularization only (SD), Winograd-domain regularization only (WD), and both regularizations (WD+SD).
%The accuracy is evaluated using (1) spatial-domain convolution and (2) Winograd convolution,\footnote{We used https://github.com/IntelLabs/SkimCaffe~\cite{li2017enabling} for Winograd convolution in accuracy evaluation.} for convolutional layers of $3\times3$ filters. In case of (2), the $3\times3$ filters are transformed into the Winograd domain and pruned to the desired ratio.
As expected, the proposed regularization method produces its desired sparsity only in the regularized domain. If we prune weights in the other domain, then we suffer from considerable accuracy loss. Using both Winograd-domain and spatial-domain regularizers, we can produce one model that can be sparse and accurate in both domains. %We can reduce the number of MACs by $2.6\times$ and $4.5\times$ when using sparse convolution in the spatial and the Winograd domains, respectively, at accuracy loss less than 0.5\%.
In Figure~\ref{sec:exp:resnet18:fig:01}, we compare the accuracy of our pruned ResNet-18 models to the ones from \cite{liu2018efficient}. %Observe that our pruned ResNet-18 models outperform the ones from \cite{liu2018efficient} in the Winograd domain. %We emphasize that the major advantage of our scheme is that it produces one model that can use any of sparse spatial-domain convolution or sparse Winograd convolution.

\begin{table}[t]
\centering
\caption{Compression results for ResNet-18 and AlexNet.\label{sec:exp:resnet18:tbl:02}}\vspace{.5em}
{\scriptsize
\begin{tabular}{ccccccc}
\toprule
Model & Method & CR &
\multicolumn{2}{c}{(1) Spatial domain} & \multicolumn{2}{c}{(2) Winograd domain} \\
\cmidrule(l{2pt}r{2pt}){4-5} \cmidrule(l{2pt}r{2pt}){6-7}
 & & & Top-1 / Top-5 & \# MACs   & Top-1 / Top-5 & \# MACs   \\
 & & & accuracy      & per image & accuracy      & per image \\
\midrule
\multirow{2}{*}{ResNet-18}
 & Pre-trained
               & -    & 68.2 / 88.6 & 2347.1M & 68.2 / 88.6 & 1174.0M \\
 & Ours        & 24.2 & 67.4 / 88.2 & 888.6M  & 67.4 / 88.2 &  516.4M \\
\midrule
\multirow{5}{*}{AlexNet}
 & Pre-trained
               & -    & 56.8 / 80.0 & 724.4M  & 56.8 / 80.0 & 330.0M \\
 & Ours        & 47.7 & 56.1 / 79.3 & 240.0M  & 56.0 / 79.3 & 142.6M \\
\cmidrule{2-7}
 & Han et al.~\cite{han2015deep}
               & 35.0 & 57.2 / 80.3 & 301.1M  & N/A         & N/A    \\
 & Guo et al.~\cite{guo2016dynamic}
               & N/A  & 56.9 / 80.0 & 254.2M  & N/A         & N/A    \\
 & Li et al.~\cite{li2017enabling}
               & N/A  & N/A         & N/A     & 57.3 / N/A  & 319.8M \\
\bottomrule
\end{tabular}
}
\end{table}

Table~\ref{sec:exp:resnet18:tbl:02} shows the compression results for the ResNet-18 and AlexNet models. %We take the model obtained by training with both Winograd-domain and spatial-domain regularizers for the desired sparsity level of 80\% and compress its weights as described in Section~\ref{sec:univ}. %We compare uniform quantization (UQ) and dithered uniform quantization (DUQ). 
%We used \textit{bzip2} as our universal source coding scheme.
The compression ratio (CR) is the ratio of the original model size to the compressed model size. 
%The results show that we achieve more than $24\times$ and $47\times$ compression for ResNet-18 and AlexNet, respectively, at accuracy loss less than 1\% in both domains.
%
%Table~\ref{sec:exp:resnet18:tbl:02} also includes the compression results for AlexNet. The proposed method produces a $47.7\times$ compressed AlexNet model at accuracy loss less than 1\% in both domains. %We compare our results to \cite{han2015deep,guo2016dynamic} in the spatial domain and to \cite{li2017enabling} in the Winograd domain.
We note that the previous approaches~\cite{han2015deep,guo2016dynamic,li2017enabling} produce sparse models only in one domain, while our method produces one compressed model that can be used in both domains.

\section{Conclusion} \label{sec:conclusion}

We introduced a CNN pruning and compression framework for hardware and/or software platform independent deployment. The proposed scheme produces one compressed model whose convolutional filters can be made sparse in both Winograd and spatial domains without further training. %Thus, one compressed model can be deployed on any platform and the sparsity of its convolutional filters can be utilized for complexity reduction in either domain, unlike the previous approaches that yield sparse models in one domain only.
We showed that the proposed method successfully compresses ResNet-18 and AlexNet with compression ratios of $24.2\times$ and $47.7\times$, while reducing their complexity by $4.5\times$ and $5.1\times$, respectively, when using sparse Winograd convolution. Our regularization method can be extended for sparse frequency-domain convolution, which remains as our future work. %It will be also interesting to compare our partial L2 norm to $k$-support norm~\cite{argyriou2012sparse} for sparsity regularization.

{\small
\newcommand{\BIBdecl}{\setlength{\itemsep}{0.25em}}
\bibliographystyle{IEEEbib}
\bibliography{manuscript}
}
%\bibliographystyle{IEEEbib}
%\bibliography{ref}

\end{document}